# Classification of Log-Polar-Visual Eigenfaces using Multilayer Perceptron

Mrinal Kanti Bhowmik, Debotosh Bhattacharjee, Mita Nasipuri, Mahantapas Kundu, Dipak Kumar Basu

**Abstract** In this paper we present a simple novel approach to tackle the challenges of scaling and rotation of face images in face recognition. The proposed approach registers the training and testing visual face images by log-polar transformation, which is capable to handle complicacies introduced by scaling and rotation. Log-polar images are projected into eigenspace and finally classified using an improved multi-layer perceptron. In the experiments we have used ORL face database and Object Tracking and Classification Beyond Visible Spectrum (OTCBVS) database for visual face images. Experimental results show that the proposed approach significantly improves the recognition performances from visual to log-polar-visual face images. In case of ORL face database, recognition rate for visual face images is 89.5% and that is increased to 97.5% for log-polar-visual face images whereas for OTCBVS face database recognition rate for visual images is 87.84% and 96.36% for log-polar-visual face images.

## 1 Introduction

Face recognition has many practical applications, such as bankcard identification, access control, mug shots searching, security monitoring, and surveillance systems[7, 16, 28]. Face recognition is used to identify one or more persons from still image or a video image sequence of a scene by comparing input images with faces stored in a database. Face recognition has the benefit of being a passive, non-intrusive system, which can verify personal identity without the consent of the concerned person or individual. Even though humans can detect and identify faces in a scene with little or no effort, building an automated system that accomplishes such objectives is very challenging, The challenges are even more profound when one considers the large variations in the visual stimulus due to illumination conditions, viewing directions or poses, facial expressions, aging, and disguises such as facial hair, glasses, or cosmetics. To address these issues several attempts have been made to increase the performance of face recognition.

Mrinal Kanti Bhowmik
Department of Computer Sc. & Engg., Tripura University, Suryamaninagar, Tripura- 799130,
email: mkb_cse@yahoo.co.in
Debotosh Bhattacharjee
Department of Computer Sc. & Engg, Jadavpur University, Kolkata-700032,
email: debotosh@indiatimes.com
Mita Nasipuri
Department of Computer Sc. & Engg, Jadavpur University, Kolkata-700032,
email: mita_nasipuri@gmail.com
Mahantapas Kundu
Department of Computer Sc. & Engg, Jadavpur University, Kolkata-700032,
email: mkundu@icse.jdvu.ac.in
Dipak Kumar Basu
AICTE Emeritus Fellow, Department of Computer Sc. & Engg, Jadavpur University,
Kolkata-700032, email: dipakkbasu@gmail.com

Mrinal Kanti Bhowmik, Debotosh Bhattacharjee, Mita Nasipuri, Mahantapas Kundu, Dipak Kumar Basu

The main objective of this work is to improve the performance of the face recognition system subject to following variations:

● **Scale invariance:** The same face can be presented to the system at different scales. This may happen due to the varying distance between the face and the camera. As this distance gets closer, the face image gets bigger.

●**Pose invariance:** The same face can be presented to the system at different perspectives and orientations. For instance, pose of the face images of the same person may appear different due to rotation and tilting.

●**Emotional expression and detail invariance:** Face images of the same person can differ when smiling or laughing and with some details such as dark glasses; beards or moustaches can be present.

Recognition performed by the human being can be simultaneously seen as a holistic and a feature analysis approach [2]. Automatic face recognition often favors only one of these aspects. Features used for description of faces are either biometric features of the face, like distances between parts of the face like nose and mouth, or more abstract features, like filter responses on a grid [10]. Template-based methods that attempt to match well-defined portions of the face (eye, mouth) belong to the analysis category [1, 26]. The Principal Component Approach (PCA) [4, 19, 22, 24] describes images in terms of linear combinations of basis images, and thus represents a global holistic approach [3]. But results show that PCA based approaches are poor in handling variations in scale, rotation, illumination and facial hair like beard and moustache.

Recently, researchers have used log-polar transform [18, 20, 21] for different image analysis purposes. In this work, at first log-polar domain conversion of visual images is done, after that using these transformed images eigenfaces are computed and finally those eigenfaces thus found are classified using a multilayer perceptron.

The organization of the rest of this paper is as follows:

In section II, the overview of the system is discussed, in section III experimental results and discussions are given. Finally, section IV concludes this work.

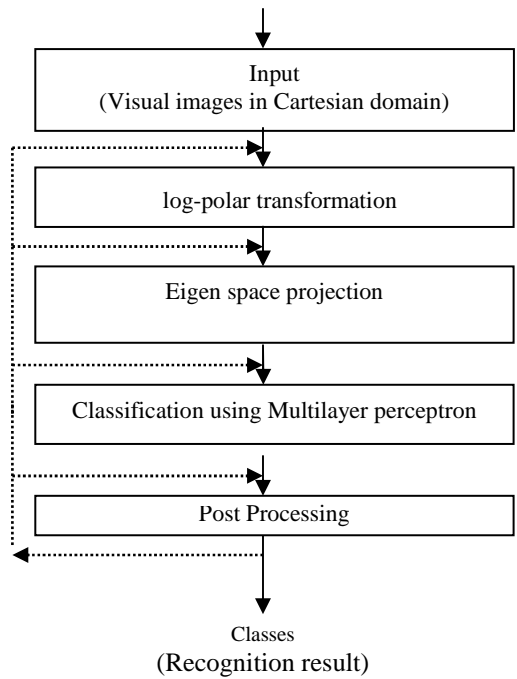

Figure 1. Block diagram of the system presented here.

**2 The System Overview**

Here we present a technique for human face recognition. In this work we have used face images from ORL face database. As this database contains faces with less rotation and no scaling, we have used Object Tracking and Classification Beyond Visible Spectrum (OTCBVS) database for visual face images, which contains faces with large rotational angle. Images are deliberately scaled up to verify the performance of the present method. In this work every face image is first converted into log-polar domain. These transformed images are separated into two groups namely training set and testing set. The eigenspace is computed using training images. All the training images and testing images are projected into the created eigenspace and named as log-polar-visual eigenfaces. Once these conversions are done the next task is to use a classifier to classify them. A multilayer perceptron is used for this purpose. The block diagram of the system is given in figure 1. In this figure dotted line indicates feedback from different steps to their previous steps to improve the efficiency of the system.

*2.1 Log-polar transformation*

The log-polar transformation is used to get rid of the problems of rotation and scaling. This transformation maps visual faces of size $M \times N$ into a new log-polar visual face image of size $Z^q \times Z^q$, Z and q will be explained subsequently. Figure 2 shows that the rotation of faces in different angles appears just column shifted in log-polar domain. Scaling has got no effect if we use a fixed size for all the images in the log-polar domain, which can be easily understood from figure 3. The Log-polar transformation algorithm is described subsequently.

**Algorithm 1: Log-polar transformation**

Input: An image of size $M \times N$ in Cartesian coordinate space.
Output: An image of size $Z^q \times Z^q$ in Log-polar coordinate space.

Step 1: For given input image of size $M \times N$, find the center (m, n) and radius (R) ensuring that the maximum number of pixels is included within the reference circle of the conversion. Center of the circle can be given as

$$m = \lfloor M/2 \rfloor, n = \lfloor N/2 \rfloor \qquad (1)$$

Step 2: Compute polar images

The pixel in the input image $(x_i, y_i)$ will be the pixel at $(r, \theta)$ position in the polar image, where

$$r = \sqrt{(x-m)^2 + (y-n)^2} \qquad 0 \leq r \leq R \qquad (2)$$

$$\theta = \tan^{-1}\left(\frac{y-n}{x-m}\right) \qquad 0 \leq \theta \leq 360^0 \qquad (3)$$

Step 3: Log-polar transform

Log-polar transform can be given as $(p, \theta)$, where $p = \log_e r$.

Step 4: Resize the image obtained in step 3 into a square image of size $Z^q \times Z^q$, where $q = \lceil \log_Z R \rceil$.

Mrinal Kanti Bhowmik, Debotosh Bhattacharjee, Mita Nasipuri, Mahantapas Kundu, Dipak Kumar Basu

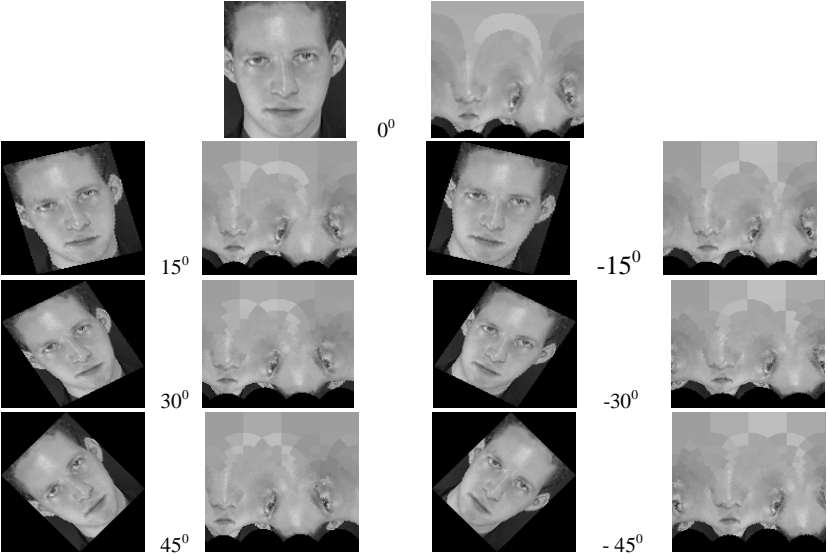

Figure 2: The Log-polar transformation for sample face image from ORL face database in rotation angles 0, +15, -15, +30, -30, +45, -45 degrees.

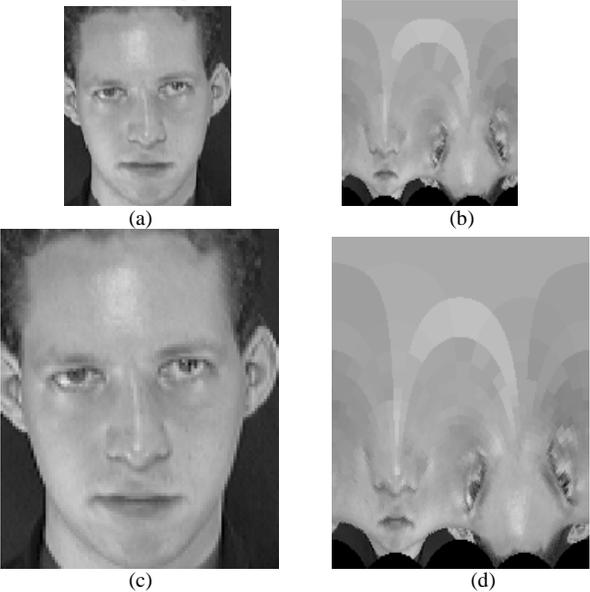

Figure 3: The Log-polar transformation of sample visual face image (first image of the figure 2.) in the scale of twice (a), (b) and thrice (c), (d) respectively.

Since sharp boundaries are not very much useful feature in case of face recognition, nearest neighbor interpolation have been used during resizing of images. Application of this algorithm in a visual face image is
shown in figure 4.

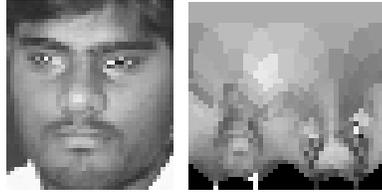

Figure 4. Conversion of one OTCBVS visual face image from Cartesian space into log-polar coordinate
space database

## 2.2 Eigenfaces for Recognition

In the context of information theory, for face recognition our aim is to extract the relevant information from a face image, encode it as efficiently as possible, and compare one face coding with the similarly encoded database contents. A simple approach is to extract the information contained in a face images, independent of any judgment of features, and use this information to encode and compare individual face images. In order to represent an arbitrary image in the image space, every pixel value is to be specified. Thus the minimum dimensionality of the space is given by the total number of pixels, which is very high number even for images of moderate size. Recognition methods that operate on such a representation must suffer from a number of potential disadvantages, which is due to curse of dimensionality:

- Recognition of high-dimensional examples using similarity-based or matching-based approaches is computationally expensive.
- For parametric methods the number of parameters, one need to estimate, typically grows exponentially with the dimensionality. Often this number is much higher than the number of images available for training, making the estimation task in the image space ill-posed.
- Similarly, for non-parametric methods, the sample complexity, which is given as the number of examples needed to efficiently represent the underlying distribution of the data, is prohibitively high.

However, major area of the 2D face surface appears to be smooth. Also, the value of a pixel is typically highly correlated with the values of the surrounding pixels. Therefore, a huge number of pixel points in the image space do not represent any crucial discriminative information. Hence, the same face images can be efficiently represented with fewer points in a subspace called face space. It is common to model the face space as a (possibly disconnected) principal manifold, embedded in the high-dimensional image space. Its intrinsic dimensionality is determined by the number of degrees of freedom within the face space. The goal of the subspace analysis is to determine this dimensionality and to extract the principal components of the subspace.

In mathematical terms, we wish to find principal components [4, 19, 22, 24] of the distribution of faces, or the eigenvectors of the covariance matrix of the set of face images. These eigenvectors can be thought of as a set of features which together characterize the variations between face images. Each image location contributes more or less to each eigenvector, so that we can display the eigenvector as a sort of ghostly face which we call an eigenface. Each face image in the training set can be presented exactly in terms of a linear combination of the eigenfaces. The number of a possible eigenfaces is equal to the number of face images in the training set. However the faces can also be approximated using only the "best" eigenfaces- those that have the largest eigenvalues, and which therefore account for the most variance within the set face images. The best U eigenfaces constitute a U-dimensional subspace, which may be called as "face space" of all possible images. Identifying images through eigenspace projection takes three basic steps. First the eigenspace must be created using training images. After that all those training images are projected into the eigenspace and call them eigenfaces. Train a classifier using these eigenfaces. Finally, the test images are

Mrinal Kanti Bhowmik, Debotosh Bhattacharjee, Mita Nasipuri, Mahantapas Kundu, Dipak Kumar Basu

identified by projecting them into the eigenspace and classifying them by the trained classifier. The eigenspace projection algorithm is given below.

**Algorithm 2: Eigenface projection**

Specifically, each image is stored in a column vector form of size H, where H= M × N.

$$X^{(i)} = [\ x_1^{(i)}\ \ x_2^{(i)}\ \ldots\ldots\ldots..x_H^{(i)}\ ]^T \qquad (4)$$

Step 1: Creation of Eigenspace

The following steps are used to create an eigenspace.

Step 1.1. Mean Centering of Data:

From each of the training images the mean image is subtracted to get centered images as shown in equation (5).

$$\overline{X}^{(i)} = X^{(i)} - m \qquad (5)$$

where $m = \dfrac{1}{p}\sum_{j=1}^{p} X^{(j)}$

Step 1.2. Formation of data Matrix

Centered images are combined column-wise into a matrix of size $H \times P$, where $P$ is the number of training images shown in equation (6).

$$X = \begin{bmatrix} \overline{X}^{(1)} & \overline{X}^{(2)} & . & . & \overline{X}^{(P)} \end{bmatrix} \qquad (6)$$

Step 1.3 Compute Covariance Matrix

Covariance matrix is created by multiplying the data matrix by its transpose, shown in equation (7).

$$\Omega = X \cdot X^T \qquad (7)$$

Step 1.4. Eigenvalues and Eigenvectors calculation

For the covariance matrix, the eigenvalues and corresponding eigenvectors are computed as shown in equation 8.

$$\Omega V = AV \qquad (8)$$

Here, $V$ is the collection of eigenvectors for which corresponding eigenvalues are given in A.

Step 2. Creation of Eigenspace

At first, all the eigenvectors for which eigenvalue is not greater than zero are deleted. Let us consider the number of such eigenvector remains is U. So, the collection of those eigenvectors column-wise gives us the eigenspace $V$, shown in equation 9.

$$V = \begin{bmatrix} v_{11} & v_{12} & . & . & v_{1U} \\ v_{21} & v_{22} & . & . & v_{2U} \\ . & . & . & . & . \\ . & . & . & . & . \\ v_{H1} & v_{H2} & . & . & v_{HU} \end{bmatrix} \qquad (9)$$

Step 3. Projection of Images

All the images irrespective of training or testing needed to be projected into the eigenspace, which can be done by equation (10). Each of the centered training images is projected into the eigenspace. To project an image into the eigenspace, calculate the dot product of the image with each of the ordered eigenvector.

$$\tilde{X}^{(i)} = V^T \cdot \overline{X}^{(i)} \qquad (10)$$

where, $\overline{X}^{(i)}$ and $\tilde{X}^{(i)}$ are centered and projected images respectively. Projected images are called as eigenfaces.

## 2.3 *ANN using Backpropagation with Momentum*

Neural networks, with their remarkable ability to derive meaning from complicated or imprecise data, can be used to extract patterns and detect trends that are too complex to be noticed by either humans or other computer techniques. A trained neural network can be thought of as an "expert" in the category of information it has been given to analyze. The Back propagation learning algorithm is one of the most historical developments in Neural Networks. It has reawakened the scientific and engineering community to the modeling and processing of many quantitative phenomena using neural networks. This learning algorithm is applied to multilayer feed forward networks consisting of processing elements with continuous differentiable activation functions. Such networks associated with the back propagation learning algorithm are also called back propagation networks. For the completeness of the paper a brief overview of the backpropagation learning is given as algorithm 3.

**Algorithm 3: Backpropagation learning**

Input: A set of training pairs {( $\tilde{X}^{(k)}$, $d^{(k)}$) |k=1,2,…,P},    where, $d^{(k)}$ is the desired vector representing the class to which $\tilde{X}^{(k)}$ belongs.
Output: Weight vector by which the desired classification is achieved.
Step 0: (Initialization) Choose η>0 and $E_{max}$, set E=0 and k=1.
Step 1: (Training Loop) Apply $k^{th}$ input pattern to the input Layer.
Step 2: (Forward Propagation) Propagate the Signal forward through the Network.
Step 3: (Output error Measure) Compute the error value $E = 1/2 \sum (d_i^{(K)} - y_i^{(K)})^2 + E$.
Step 4: (Error back propagation) Propagate the errors backward to update weights.
Step 5: (One epoch Looping) If k < P then k=k+1, go to Step 1.
Step 6: (Total error checking) Check whether the current total error is acceptable. If $E < E_{max}$ then terminate the training process and output the final weights, otherwise E=0, k=1 and go to Step 1.

The main disadvantage of the back-propagation algorithm is its slow convergence. This is due to the reason that the error surfaces have a multitude of areas with shallow slopes in multiple dimensions. This usually happens because the outputs of some nodes are large and insensitive to small weight changes and take place on the shallow tails of the sigmoid function. Due to the stochastic nature of the algorithm, the randomness helps it to get out of local minima but the stochastic nature provides an instantaneous estimation of the gradient of the error surface in weight space. When the error surface is almost flat along some weight dimension, the derivative of the error surface with respect to that weight direction is close to zero in magnitude. Due to that, the weight updating is very small and consequently less error reduction in individual iterations, which leads to slow convergence i.e. a large number of iterations for learning. If that updating amount is increased by larger learning parameter then in the case of curved error surface along a weight dimension the derivative would be large in magnitude, which may overshoot the minimum point of error surface or the learning process may oscillate about that minimum point. To overcome the difficulty that may arise from the slow convergence of the Backpropagation algorithm, following rules are considered to accelerate the algorithm's convergence.

i)  Instead of a constant learning rate parameter, it may vary from one iteration to other,
ii) If the derivative of the error function with respect to some weight appears with same algebraic sign consistently the learning-rate parameter for that weight dimension should be increased and,
iii) If the algebraic sign of the derivative of the error function with respect to a particular weight dimension oscillates for several consecutive iterations, the learning-rate parameter for that weight direction should also be increased.

Mrinal Kanti Bhowmik, Debotosh Bhattacharjee, Mita Nasipuri, Mahantapas Kundu, Dipak Kumar Basu

All these rules are incorporated in the algorithm called Delta-Bar-Delta Learning Algorithm [14] with
momentum ($\alpha$). In this method the weight adjustment is done by the equation (11), which is given below.

$$\Delta w(t) = -\eta \nabla E(t) + \alpha \Delta w(t-1), \quad where \quad , \quad \alpha \in [0,1] \tag{11}$$

The value for the momentum term has been used as 0.9. If $\alpha=0$ this becomes original backpropagation algorithm and with larger value of $\alpha$, each weight change is enhanced by the contribution of the downhill force from the previous time step.

The learning-rate parameter adjustment is done by equation (12) given below. (12)

$$\Delta\eta(t+1) = \begin{cases} +a & if \ \bar{\lambda}(t-1)\lambda > 0 \\ -b\eta & if \ \bar{\lambda}(t-1)\lambda < 0 \\ 0 & otherwise \end{cases} \quad where \quad \lambda(t) = \frac{\partial E}{\partial w_{ij}} \quad and \quad \bar{\lambda}(t) = (1-c)\lambda(t) + c\bar{\lambda}(t-1)$$

where, $c \in [1, 0]$.

If the parameters a and b in equation (12) are set to zero, then the learning rate parameters take a constant value, as in the original back-propagation algorithm. Here, the learning-rate parameters adjustment is obtained as linearly increasing and exponentially decreasing. The linear increase does not allow fast increase rates, whereas the exponential decrease denotes that the learning-rate parameters remain positive and are decreased fast.

## 3 Experiment Results and Discussions

For comparison of results experiments are conducted for visual as well as log-polar-visual images. A thorough system performance investigation, which covers two specific conditions of human face recognition, has been conducted. They are face recognition under
      i) Variations in size.
      ii) Variations in pose.

### *3.1 Face database*
We analyze the performance of our algorithm using ORL face database and OTCBVS database for visual face images.

### 3.1.1 ORL face database
Cambridge (ORL) database [5] presently AT & T Cambridge face database, containing 400 images of 40 distinct persons with variation in expression, illumination, pose etc.

### 3.1.2 OTCBVS face database
Object Tracking and Classification Beyond Visible spectrum (OTCBVS) benchmark database [6] contains a set of visual face images. There are 2000 visual images of 16 different persons. For some subject, the images were
taken at different times which contain quite a high degree of variability in lighting, facial expression (open / closed eyes, smiling /non smiling etc.), pose (Up right, frontal position etc.) and facial details (Glasses/ no Glasses). All the images were taken against a dark homogeneous background with the subjects in fontal position, with tolerance for some tilting and rotation of up to 20 degree. Some sample images and their corresponding log-polar transforms are shown in figure 5.

## *3.2 Classification of Log-polar-visual Eigenfaces using Multilayer Perceptron*

In case of ORL face database 200 images are used as training images and 200 images used as Testing images. Out of total 2000 visual images from OTCBVS face database 1120 images are used as training set and rest 880 images are taken as testing images. All these visual images are first transformed into log-polar domain. In this work a multilayer neural network with backpropagation has been used. The learning algorithm error back propagation with momentum is used here. Momentum allows the network to respond not only to the local gradient, but also to recent trends in the error surface. We have used momentum to back propagation learning by making weight changes equal to the sum of a fraction of the last weight change and the new change suggested by the back propagation rule. The gradient is computed by summing the gradients calculated at each training example, and the weights and biases are only updated after all training examples have been
presented.

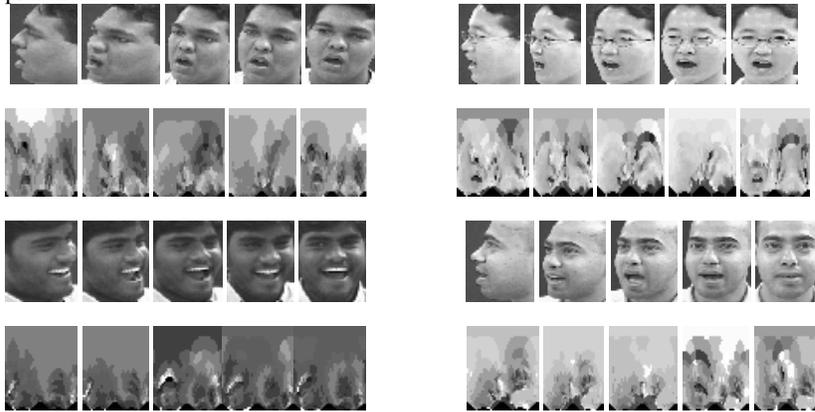

Figure 5. Some samples from OTCBVS face database and corresponding log-polar transforms given below.

The design of a neural network is not very easy. There are two major approaches to finalize the number of hidden layers and number of nodes in each of the hidden layers of a network [9]. The first method is called as pruning algorithm where the training process starts with a larger network so that an acceptable solution is found. After that, some hidden units are removed without degrading the performance of the network. The second method is called constructive approach, which starts with a small network and then grows with additional hidden nodes and layers to find acceptable solution. Constructive algorithm is
straightforward and also finds a smaller network in comparison to pruning. In case of pruning it is difficult to find the initial network, which can classify the given inputs successfully. Therefore, it is better to use constructive approach. Moreover, in this work we have taken constructive approach [9], keeping following two contradictory requirements:

       (i) Faster convergence of the network.
       (ii) Better classification performance.

Mrinal Kanti Bhowmik, Debotosh Bhattacharjee, Mita Nasipuri, Mahantapas Kundu, Dipak Kumar Basu

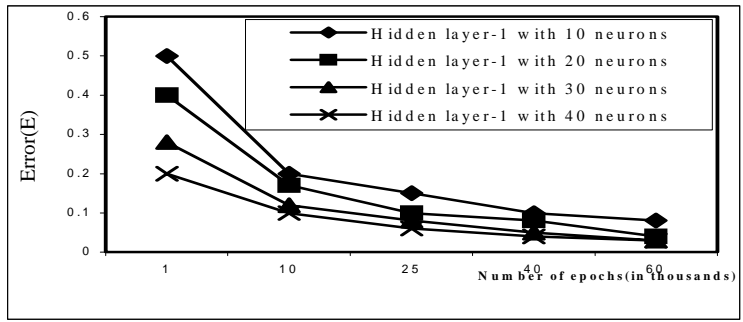

Figure 6. Plot for total error vs. no. of epochs for different number of neurons in hidden layer-1

Initially, experiments were conducted without any hidden layer, but the network did not converge. Therefore, hidden layers were introduced with an assumption that the given feature vectors are linearly non-separable. To decide about the number of neurons in hidden layer, several training experiments were conducted. Along with these examples (positive) all these training experiments were conducted with five other classes as negative examples. A trend of decrease in total error (E) with increase in the size of hidden layer-1 has been observed, as shown in Figure 6.

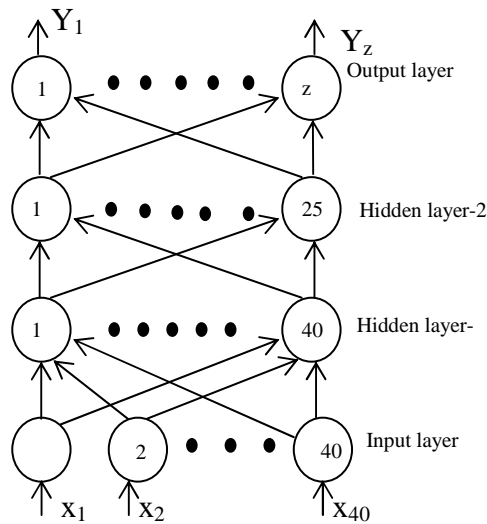

Figure 7. Architecture of the MLP used in this work.

There is no significant improvement in error minimization if the number of neurons is increased beyond 40. Although there is no harm in keeping more hidden nodes, but it incurs unnecessary wastage of time and space to store more number of nodes and interconnections due to them. So, the size of the hidden layer-1 is taken as 40 neurons. Similar experiments have been conducted for the hidden layer-2 the size of that have been finalized as 25 as shown in figure 7. In the input layer number of nodes should be U as per equation (9), but that varies for input images. To fit that to an MLP number of eigenvectors is kept to be constant as 40, which is also the size of the feature or input vector to the MLP and hence the size of the input layer of the

Classification of Log-polar-Visual Eigenfaces using Multilayer Perceptron

MLP is 40. Since total number of persons in ORL and OTCBVS are 40 and 16 respectively, the size of the output layer is also kept as 40 and 16 respectively for those databases.

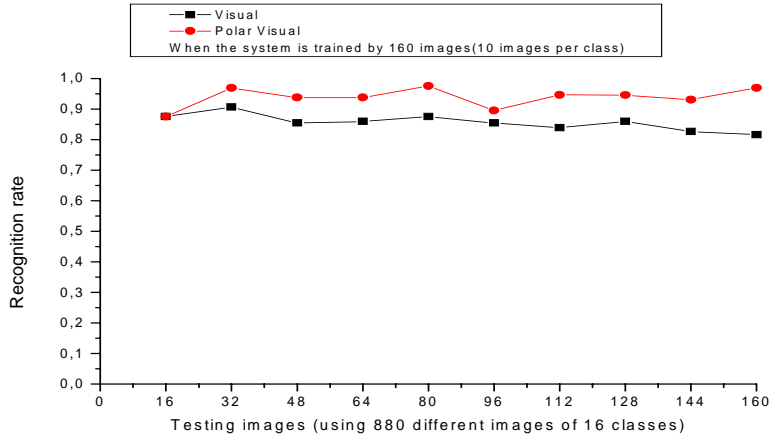

Figure 8. Comparative study of recognition rate of visual and log-polar-visual face images.

For this work the application software environment was chosen is MATLAB 7. The functions and parameters used for implementing this network are described below:

i) *newff*: It is the function that creates a feed forward network. It requires four inputs:
    a) The first input is an R by 2 matrix of minimum and maximum values of each of the R elements of input vectors, which is done by *minmax* function.
    b) An array containing the size of each layer.
    c) Third input is a cell array containing the names of the transfer functions to be used in each layer. Here, *tansig* function has been used. It is a tansigmoid transfer function used in multilayer network that produce output between 1 and -1.
    d) A training strategy for the network. Here, we have used *traingdm*, which is a network training function with a momentum that updates weight and bias values according to gradient descent.

iv) *train*: It is a function used to train the network and it takes three arguments, the network created by *newff*, the input vector, and a vector that contains desired outputs.

v) Different parameters to train the network:
    a) *epochs*: Number of iteration to train the network, which has been considered here as 70000. This is used to stop the training process in finite number of iterations. Otherwise, on the occasion of non-convergence of the network the training process would proceed continuously for infinite times.
    b) *goal*: Performance goal. The training stops when performance goal is met. Here it is given in terms of change in gradient and the value is chosen as $10^{-6}$.
    c) *lr*: Learning rate, which should be moderately small, taken as 0.02.
    d) *mc*: The magnitude of the effect that the last weight change is allowed to have is mediated by a momentum constant (*mc*). As discussed earlier in the subsection 2.3 this may be any number between 0 and 1. Here, we have considered $mc = 0..9$.

Mrinal Kanti Bhowmik, Debotosh Bhattacharjee, Mita Nasipuri, Mahantapas Kundu, Dipak Kumar Basu

In order to assess the effectiveness of log-polar-visual images we compare face images from visual spectra. Results obtained after applying the same procedure for visual and log-polar-visual images are shown in figure 8. Here, experiments are conducted for different number of images during testing. Trend of false rejection error for different test images is depicted in figure 9.

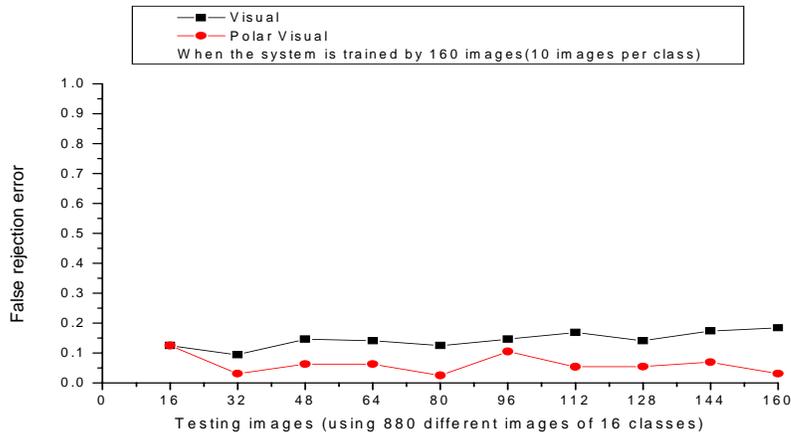

Figure 9. Comparative study of false rejection rate of visual and log-polar-visual face images.

Comparison of error rates for ORL face database for the present method and other methods are shown in table 1. In this table, face recognition with RBF neural network [8], has shown better performance than the present method for the same ORL face database, but at the cost of huge computational effort. This is due to the reason that the method in [8] first extracts face features by PCA and then the resulting features are projected into the Fisher's optimal subspace and finally, a hybrid learning algorithm is proposed to train the RBF neural networks for classification.

**Table 1. Comparison of error rate with other Methods**

| Face recognition method | | Error rate (%) |
| --- | --- | --- |
| Eigenface [22] | | 10 |
| CNN [11] | | 3.8 |
| PDBNN [13] | | 4 |
| NFL [12] | | 3.125 |
| PCA + RBFN [8] | | 4.9 |
| Wavelet + RBFN [8] | | 3.7 |
| PCA+Fisher+ RBFN [8] | | 1.92 |
| Kernel Asso. Mem. [27] | | 3.7 |
| **Present method** | **using visual images** | **10.5** |
| | **using log-polar-visual images** | **2.5** |

Comparison of recognition rates for the present method using OTCBVS and other commonly referred methods are shown in Table 2 for a quick comparison. Although they use different face databases, i.e. other than face database, the present method can be compared favorably against other relevant face recognition methods.

**Table 2. Comparison of recognition rate with other Methods**

| Method | | Recognition rate (%) |
|---|---|---|
| **Present method (OTCBVS)** | using visual images | 87.84 |
| | using log-polar-visual images | 96.36 |
| InfoGabor-GDA[16] | | 95.0 |
| PCA [16] | | 83.4 |
| SQI [19] | | 92.0 |
| Wavelet + RBF [17] | | 96.3 |
| Wavelet Subband + Kernel associative Memory [18] | | 84.0 |

## 4 Conclusions

In this paper we have presented log-polar-visual face recognition results in varying ndscale, facial expression, pose, and facial details. The scheme described here performs face recognition by combining the techniques of log-polar transformation, eigenspace projection, and classification using multilayer perceptron. Eigenspace projection has been advocated by the pattern recognition community for a long time for a broad range of applications. The efficiency of our scheme has been demonstrated on ORL face database, which has shown 89.5% and 97.5% recognition rates for visual and log-polar-visual face images respectively. Also, the present technique is applied on OTCBVS database and recognition rates for visual and log-polar-visual face images are obtained as 87.84% and 96.36% respectively. This enhancement in the performance of the recognition scheme is due to rotation and scale invariant representation of face images in log-polar domain. This scheme may also be used for other types of pattern recognition and computer vision applications.

## Acknowledgment

Authors are thankful to the "Centre for Microprocessor Application for Training Education and Research" at the Department of Computer Science and Engineering, Jadavpur University, Kolkata - 700 032 and Department of Computer Science and Engineering, Tripura University, Tripura, for providing the necessary facilities for carrying out this work. Second and fifth authors acknowledge with thanks the receipts of Jadavpur University Research Grant and AICTE Emeritus Fellowship (1-51/RID/EF(13)/2007-08) respectively.

Mrinal Kanti Bhowmik, Debotosh Bhattacharjee, Mita Nasipuri, Mahantapas Kundu, Dipak Kumar Basu